\documentclass[letterpaper]{article} 
\usepackage{aaai2026}  
\usepackage{times}  
\usepackage{helvet}  
\usepackage{courier}  
\usepackage[hyphens]{url}  
\usepackage{graphicx} 
\usepackage{natbib}  
\usepackage{caption} 
\usepackage{algorithm}
\usepackage{algorithmic}
\usepackage{multirow}
\usepackage{newfloat}
\usepackage{listings}
\usepackage{xcolor}
\usepackage{colortbl}
\DeclareCaptionStyle{ruled}{labelfont=normalfont,labelsep=colon,strut=off} 
\floatstyle{ruled}
\newfloat{listing}{tb}{lst}{}
\floatname{listing}{Listing}
\title{Exploiting Block Coordinate Descent for Cost-Effective LLM Model Training}
\author{
\normalfont
  Zeyu Liu\textsuperscript{2}\thanks{The two authors contribute equally to this work.},
  Yan Li\textsuperscript{4}\footnotemark[1],
  Yunquan Zhang\textsuperscript{1,2,3},
  Boyang Zhang\textsuperscript{5},
  Guoyong Jiang\textsuperscript{6}, \\
  Xin Zhang\textsuperscript{7},
  Limin Xiao\textsuperscript{4},
  Weifeng Zhang\textsuperscript{4},
  Daning Cheng\textsuperscript{3}\thanks{Corresponding author.} \\[1ex] 
  \textsuperscript{1}National Supercomputing Center in Taiyuan \\
  \textsuperscript{2}School of Computer Science and Technology, North University of China \\
  \textsuperscript{3}Institute of Computing Technology, Chinese Academy of Sciences \\
  \textsuperscript{4} Intelligent Computing Infrastructure Lab, Lenovo Research \\
  \textsuperscript{5}University of the Chinese Academy of Sciences \\
  \textsuperscript{6}State Key Laboratory of Integrated Service Networks, Xidian University \\
  \textsuperscript{7}School of Advanced Interdisciplinary Sciences \\
}
\usepackage{bibentry}

\begin{document}

\maketitle

\begin{abstract}
Training large language models typically demands extensive GPU memory and substantial financial investment, which poses a barrier for many small- to medium-sized teams. In this paper, we propose a full-parameter pre-training and fine-tuning framework based on block coordinate descent (BCD), enhanced with engineering optimizations, to enable efficient training of large-scale models on cost-effective RTX 4090, A100 and A800 GPU clusters. Under identical hardware configurations, we reduce the training cost of a 7B model to 33\% on A100/A800 and only 2.6\% on RTX 4090, compared to standard full-parameter training. It also enables large models previously restricted to A100 clusters to be trained on RTX 4090 without degrading performance. BCD achieves comparable or better accuracy than full-parameter and fine-tuning methods at most cases, with lower GPU consumption and improved hardware utilization.

\end{abstract}


\section{Introduction}
As the scale of pre-trained language models (LLMs) continues to grow, the associated GPU memory and financial costs are increasing exponentially, posing a significant barrier for small- and medium-sized research teams and enterprises. During full-parameter training, using the Adam optimizer to train 1024 tokens for a model with $W$ parameters requires storing gradients, optimizer states, and activation values, resulting in a total memory footprint of around $4.7W$. Even with the combination of recomputation techniques, the memory demand remains as high as $4W$. Furthermore, taking a 20B-parameter model as an example, the total training cost can reach the million-dollar range, severely limiting the adoption and deployment of LLM technologies.

Conventional 3D distributed parallel training mitigates single-GPU memory constraints by partitioning model parameters, gradients and optimizer states across multiple devices. However, its growing costs have become increasingly evident. As model scales expand, inter-node communication frequency and data volume rise sharply, with communication overhead consuming a significant portion of training time. Simultaneously, operating dozens to hundreds of high-end A100/A800 GPUs leads to exorbitant compute costs, substantially increasing overall financial investment. In contrast, as shown in Table \ref{4090/a100}, cost-effective yet high-performance GPUs such as RTX 4090 have become increasingly available. However, despite their advantages in compute capability and affordability, their limited memory capacity and interconnect bandwidth make large-scale training infeasible. This highlights a critical gap between hardware cost-efficiency and the scalability requirements of modern model training.

\begin{table}[ht]
\centering
\renewcommand{\arraystretch}{1.2}
\setlength{\tabcolsep}{6.0pt}
\scalebox{0.88}{
\begin{tabular}{c|cccccc}
\hline
\textbf{Feature} & \textbf{RTX 4090} & \textbf{A100} \\ \hline
\textbf{CUDA Cores}  & \textcolor{blue}{16,384}  & 6,912  \\ 
\textbf{Tensor Cores }  & \textcolor{blue}{4th Gen}  & 3rd Gen \\ 
\textbf{Memory}  & 24 GB  & \textcolor{blue}{80 GB }      \\ 
\textbf{Bandwidth} & 1,008 GB/s  & \textcolor{blue}{1,935 GB/s} \\ 
\textbf{Memory Bus}  & 384-bit  & \textcolor{blue}{5,120-bit}   \\ 
\textbf{FP32 Perf}& \textcolor{blue}{~82.6 TFLOPS  } & ~19.5 TFLOPS \\ 
\textbf{Tensorcore FP16} & \textcolor{blue}{~330 TFLOPS}  & ~312 TFLOPS \\ 
\textbf{Economic Cost} &\textcolor{blue}{0.29\$/hour} &1.20\$/hour \\ \hline
\end{tabular}}
\caption{Comparison of   RTX 4090 and NVIDIA A100. {RTX 4090's computational performance and economic cost are better then A100, but not suitable for LLM training.}  LLMs are trained in A100 for its large memory and high bandwidth but the economic cost is high.}
\label{4090/a100}
\end{table}

Despite the growing availability of low-cost GPU clusters, there is still no mature algorithm capable of fully utilizing them under resource constraints. The predominant solution, DeepSpeed Offloading, allows training on limited GPU resources but suffers from significant performance degradation and increased training time. These inefficiencies stem from frequent data transfers between GPU and host memory, such as those involving parameters, optimizer states, activations, and gradients, often without delivering a corresponding reduction in total training cost.

In this context, we propose a full-parameter pre-training paradigm based on Block Coordinate Descent (BCD), enhanced with several engineering optimizations. Specifically, BCD adopts a block-wise parameter update strategy: the full set of model parameters is partitioned into blocks according to a predefined scheme, and only one block is updated per iteration. Furthermore, parameters that are not updated in a given iteration do not require storage of optimizer states, gradients, or activations. When $1/3$ of the parameters are updated at each step, peak memory usage can be reduced to less than 50\% of conventional recomputation strategies. To ensure convergence and training stability, BCD leverages its theoretical foundations to guarantee global convergence, and employs a phased update strategy to maintain robust training dynamics.

Based on this framework (code address in Appendix) we optimize and validate the cost-efficiency of BCD under pre-training and fine-tuning scenarios: 

\begin{itemize}
\item[$\bullet$] Lower Cost on Same Hardware: For a 7B model, under identical hardware settings, BCD reduces pre-training cost to approximately 33\% on A100/A800 clusters on averagely and only 2.6\% on RTX 4090 clusters. This approach achieves substantial cost reductions compared to conventional full-parameter training methodologies.
\item[$\bullet$] Cross-Device Transfer: Through BCD technology, large-scale pre-training models that previously required expensive A100/A800 GPU clusters can now be efficiently deployed on more affordable hardware configurations (such as RTX 4090 systems). Our experimental results demonstrate that BCD reduces memory requirements while maintaining full-parameter training capability, decreasing operational costs by 75\% compared to conventional A100 clusters. This advancement significantly lowers the infrastructure barrier for large-scale model development..

\item[$\bullet$] Accuracy Retention: In both full-parameter and fine-tuning training scenarios, BCD delivers accuracy comparable to or better accuracy than traditional training methods in most case, while also exhibiting lower overall economic cost compared to distributed training methods.
\end{itemize}

\section{Related Work}
\subsection{Block Coordinate Descent}

Block Coordinate Descent is a well-established optimization method that does not rely on gradient descent, and its theoretical properties have been extensively studied and analyzed \cite{tseng2001convergence,beck2013convergence,wright2015coordinate,richtarik2014iteration,cai2023cyclic,nutini2022let,tu2016large}. In recent years, several studies have explored the application of BCD to deep neural network (DNN) training \cite{zisselman2019local,zhao2014accelerated,blondel2013block,wu2021federated,damaskinos2021differentially}.
The convergence of BCD on DNNs has been theoretically established \cite{zeng2019global,zhang2017convergent}. Some works \cite{zeng2019global,lau2018proximal} also report that in certain scenarios, BCD can outperform traditional optimizers in terms of model performance. Compared to full-parameter training, BCD generally requires more iterations but offers substantial savings in memory and computational resources. However, to the best of our knowledge, no prior work has investigated the use of BCD for training large language models (LLMs) or analyzed its potential advantages in reducing economic costs.

\subsection{LLM Training}
\subsubsection{Distributed Parallel Training}
Large-scale model training is predominantly conducted through distributed parallelism \cite{narayanan2021efficient, lai2023merak}, particularly 3D parallelism, which partitions model parameters, optimizer states, and other components across multiple GPUs.
This technique is widely adopted in advanced training frameworks, including DeepSpeed \cite{holmes2024deepspeed, aminabadi2022deepspeed, rajbhandari2022deepspeed} and Megatron-LM \cite{shoeybi2019megatron}.

\subsubsection{Offloading}
To reduce GPU memory consumption, offloading techniques dynamically transfer model parameters, gradients, and optimizer states from GPU memory to CPU memory or other storage devices \cite{rajbhandari2021zero,narayanan2021efficient,aminabadi2022deepspeed}.
It significantly enhances the feasibility of training larger model in resource-constrained environments and is widely adopted across various frameworks. However, it often leads to a substantial decrease in training speed. \cite{gao2024cost, zhang2024adam, athlur2022varuna, lv2023full}.

\section{Methodology}
\subsection{Exploit BCD for LLM Training}
The BCD method facilitates the training of large models by significantly reducing memory consumption during the training process. Consider the case of single-GPU training: when a single GPU has sufficient memory to hold the model, the amount of parameters involved in each training iteration can be dynamically adjusted based on the available memory. In the most extreme scenario, training can still proceed as long as enough memory is available to update a single parameter. For example, using a RTX 4090 with 24GB of memory, it is theoretically feasible to train a 12B-parameter large language model. However, training under such memory-constrained conditions results in extremely high time overhead.

\begin{figure}[!t]
    \centering
    \includegraphics[width=0.43\textwidth]{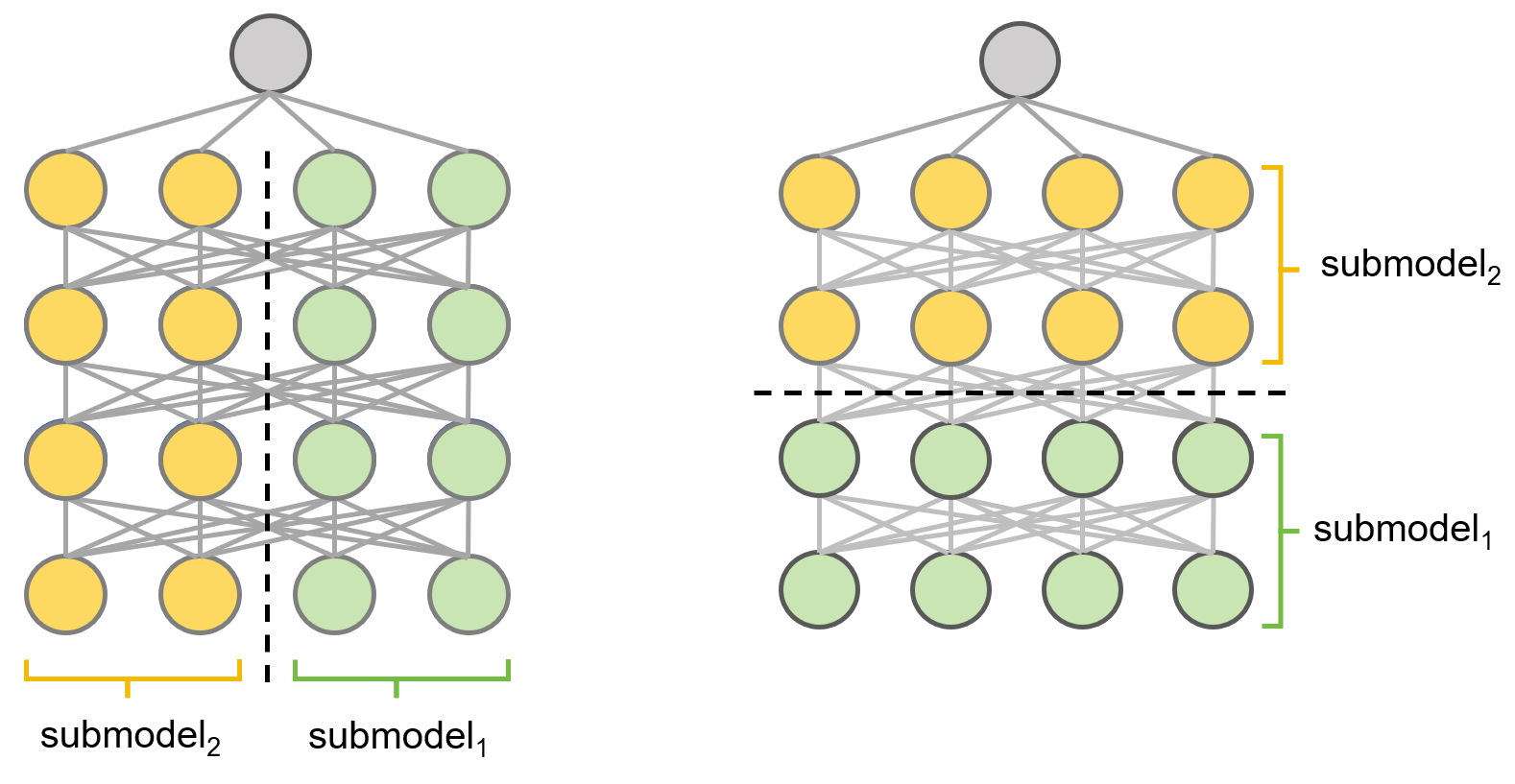}
    \caption{When we partition the model, slicing it by layers allows for better utilization of optimized computation kernels (as shown in the right Figure). If a portion of the parameters in each layer are updated simultaneously (as shown in the left Figure), computational performance decreases, and pre-inference cannot be used to accelerate the forward process.}
    \label{model cut}
\end{figure}

The standard BCD method imposes no mathematical requirements on parameter partitioning, freezing any parameters satisfies the algorithm's needs. However, in practice, the selection of which parameters to freeze should account for both the computational efficiency provided by the underlying architecture and the reduction of memory usage during training. Since the computation in neural networks typically involves a composition of optimized computational kernels, partitioning parameters based on operator boundaries is the most reasonable strategy for freezing.

In modern neural networks, most operators are encapsulated within the \texttt{layer} data structure. Therefore, freezing and unfreezing parameters at the layer-level or higher-level abstractions, such as residual blocks in ResNet, is the most appropriate approach, as illustrated in Figure \ref{model cut}. This design choice maximizes the benefits from kernel-level optimization. Accordingly, we adapt the generic BCD method to better suit the structure and computational characteristics of deep learning models, resulting in the proposed algorithm described in Algorithm~\ref{cd algorithm}.

\begin{algorithm}[!t]
\begin{algorithmic}[1]
\STATE \textbf{Input:} Training Model $model$, Traning Dataset $\mathcal{D}$,and convergence criterion.
\STATE \textbf{Initialize:} Set \( k = 0 \) and Split $model$ into $\{submodel_1,submodel_2,...,submodel_M\}$.
\REPEAT
    \STATE Select a submodel  $submodel_{i_k},i_k \in \{1, 2, \ldots, M\} $ cyclically.
    \STATE Freezing the parameters in $submodel_j,j \neq i_k$
    \STATE Unfreezing the parameters in $submodel_{i_k}$
    \STATE Building optimizer of  $submodel_{i_k}$. The opitmizer can be SGD or Adam.
    \STATE Training the $model$ on $\mathcal{D}$ until converged. Only the parameters in  $submodel_{i_k}$ is updated.
    \STATE Increment \( k \leftarrow k+1 \).
\UNTIL{Convergence criterion is satisfied.}
\STATE \textbf{Output:} Converged model $model_k$.
\end{algorithmic}
\caption{Using BCD Training LLM}
\label{cd algorithm}
\end{algorithm}

As illustrated in the algorithm, each iteration applies a standard optimizer, such as SGD or AdamW, to update the current parameter block until local convergence is reached along the given descent direction. In terms of optimizer choice, there is no fundamental constraint, as most common algorithms can effectively advance the training. Nonetheless, empirical results suggest that SGD often yields better model performance than Adam or AdamW in this setting.

\subsection{Engineering Improvements}

\subsubsection{Parallel Training}
For optimal performance in multi-GPU BCD implementations, it is essential to maintain consistently high hardware utilization across all computational devices. This requirement implies that all processing units must sustain full engagement throughout operations to achieve maximum hardware efficiency. To achieve above target, both frozen and unfrozen parameter parts should be stored simultaneously on the same hardware. While conventional parallelization strategies, such as data parallelism, tensor parallelism, and pipeline parallelism, are widely used in multi-GPU training, only data parallelism is inherently compatible with BCD. In contrast, tensor and pipeline parallelism, as forms of model parallelism, necessitate further adaptation to align with BCD’s block-wise update mechanism.

\begin{figure}[!htbp]
    \centering
    \includegraphics[width=0.45\textwidth]{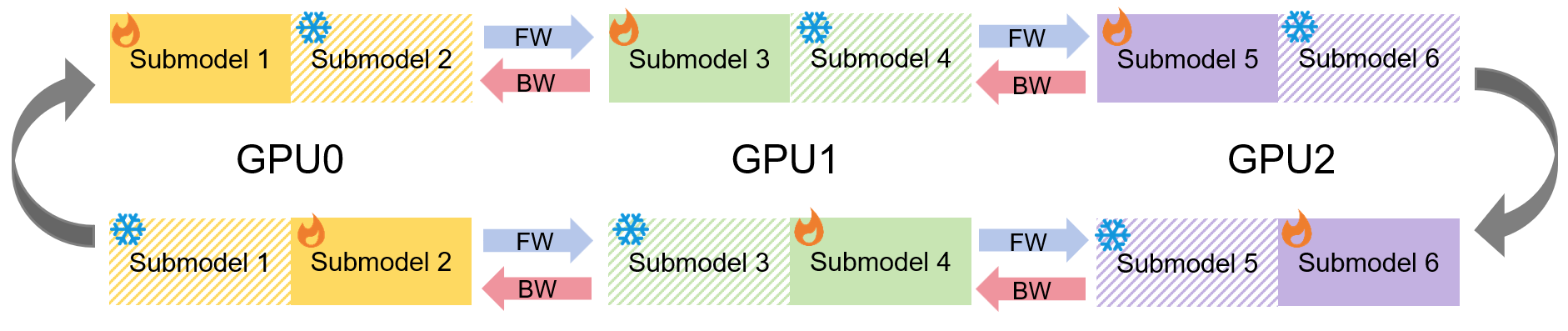}
    \caption{Training a large model on three GPUs. The model is divided into six submodels. Two submodels are addressed into one GPU. Different GPUs are paralleled via pipeline parallel manner.  The light blue part  is GPU memory.}
    \label{pipeline}
\end{figure}

Freezing and unfreezing parameters within matrix multiplication operations can substantially affect the efficiency of the underlying computation kernels. In tensor parallelism, only a portion of the matrix parameters is active during computation, which disrupts spatial locality and reduces kernel efficiency. Moreover, existing framework-level kernel optimizations are generally based on  layer-based neural network structures, further limiting the effectiveness of tensor parallelism under partial parameter activation. In addition, tensor parallelism introduces inter-device communication overhead, which increases system complexity and runtime burden. In contrast, pipeline parallelism aligns naturally with the layer-wise organization of neural networks, as illustrated in Figure~\ref{pipeline}. This alignment allows frozen layers to be skipped at the kernel level, preserving computational efficiency. As a result, pipeline parallelism is a more suitable multi-GPU strategy for BCD in large-scale model training.

\subsubsection{Pre-inference}
When the model is sufficiently partitioned, most parameters are involved only during the forward inference phase and remain frozen during backward process. Meanwhile, although the computational cost of the backward is reduced, the cost of the forward remains unchanged due to the increased number of training iterations. Therefore, in BCD methods, reducing inference time becomes necessary to achieve performance comparable to traditional full-parameter training approaches.

When training models on clusters with high computational power but limited communication bandwidth (e.g., clusters composed of RTX 4090 GPUs or Hygon DCUs), we observe that a substantial portion of computational resources often remains underutilized. Despite the high computational capacity of these clusters, their large-scale deployment often leads to costs from communication overhead, data movement, and system instability. Efficiently utilizing such clusters remains a significant challenge.

\begin{figure}[!htbp]
    \centering
    \includegraphics[width=0.43\textwidth]{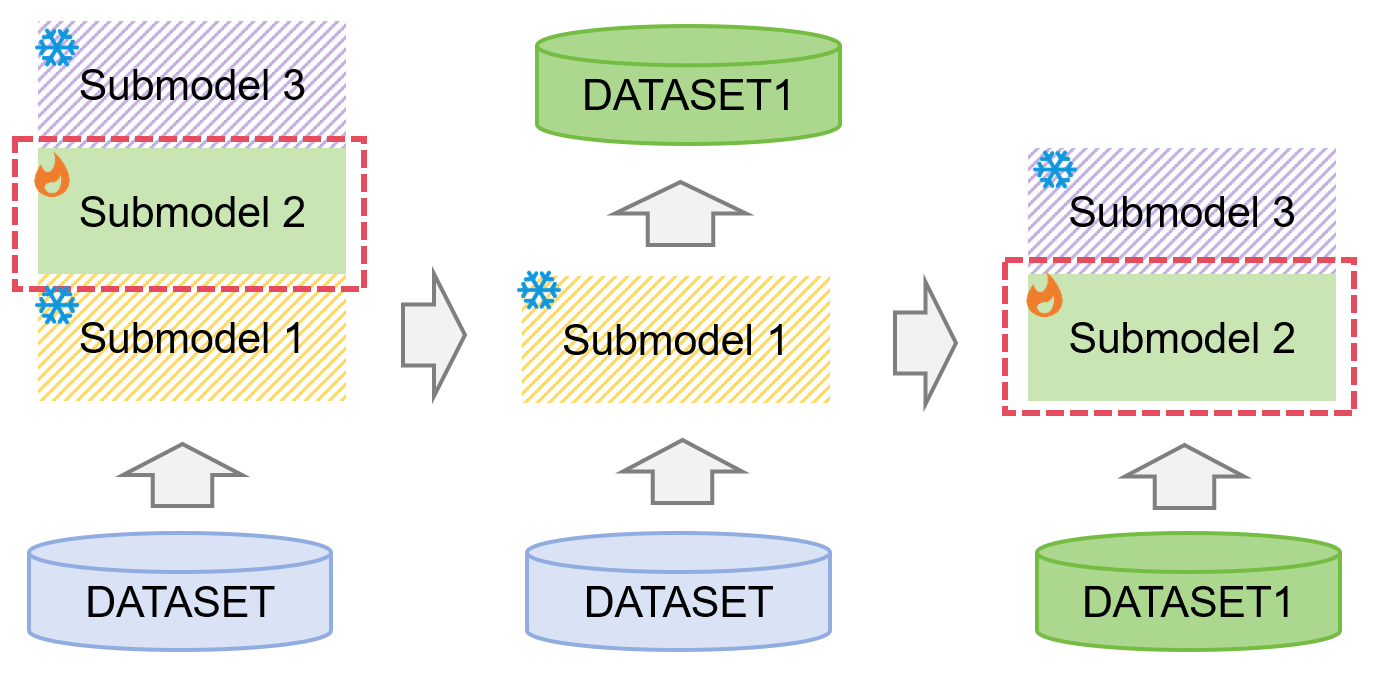}
    \caption{The process of pre-inference. We Train submodel 2 in this case. The parameters in  submodel 3 is fixed. We can inference the $Dataset$ on submodel 3 and gain $Dataset1$ on high communication cost cluster on AllReduce manner. Then Using $Dataset1$ to train the model composed by submode 1 and submodel 2. The parameters in submodel 2 is trainable. }
    \label{preinference}
\end{figure}

To address above problems, we propose a pre-inference approach in the context of model training using BCD methods to fully utilize these resources. Pre-inference refers to the situation where, when the training of submodels closer to the label, the parameters in the submodels near the data input remain unchanged. Therefore, these submodels near the data input can be treated as a complete model. In this model, all data is inferred to produce a new dataset, which is then used to train and update the parameters of the submodels,  shown in Figure \ref{preinference}.

Pre-inference is primarily aimed at accelerating computation in high-performance, high-communication-cost cluster environments. In such environments, all computational resources can be utilized for submodel inference tasks to construct new datasets, significantly reducing the overall time consumed by inference. When computational resources with high performance and high communication costs are unlimited, the inference cost of fixed submodel parts becomes a fixed time cost. This means that the time to infer a single sample is equal to the sum of the inference time and the cost of a single AllReduce operation.

\section{Experiments}
We present a systematic evaluation of the BCD framework through three  experiments: 1) Validity and Convergence: BCD achieves performance comparable to full-parameter training, with stable convergence across distributed setups, offloading configurations, and varying parameter settings. 2) Cost Efficiency: Based on results from RTX 4090, A100, and A800 platforms, we analyze training costs using cloud rental rates, showing BCD significantly reduces computational expenses. 3) Fine-tuning Performance: Compared with GaLore, and LoRA, BCD demonstrates competitive output quality and stability in downstream tasks. These results collectively confirm BCD’s advantages in accuracy, efficiency, and practicality across diverse settings.

\subsection{Experiment Setting}
In all experiments parts, all models, algorithms and framework configurations are listed in this part.

Experiments were conducted on three GPU clusters, including a single node equipped with 8 A100 or A800 GPUs, and four nodes each equipped with 8 RTX 4090 GPUs. All servers were interconnected using a 2×25Gbps high-speed network. In our experiments, we trained models based on various datasets as shown in Table~\ref{model abb} to adequately showcase the result. The implementation details and datasets are provided in Appendix.

\begin{table}[!ht]
\centering
\renewcommand{\arraystretch}{1.2}
\setlength{\tabcolsep}{6.0pt}
\scalebox{0.70}{
\begin{tabular}{c|cccccc}
\hline
{\textbf{Model}} &{\textbf{Parameters}} & {\textbf{Abbreviation}} &{\textbf{Datasets}}   \\ \hline
\multirow{8}{*}{GPT2} 
& {0.15B} &{--} & {WikiText / WebText2} \\ \cline{2-4}
& {1.5B}  &{G-1.5B} & \\ 
& {1.6B}  &{G-1.6B} & \\ 
& {2B} &{G-2B} & \multirow{2}{*}{\shortstack{WikiText\\Alpaca\\SlimPajama}} \\ 
& {5.4B} &{G-5.4B} &  \\ 
& {7B}  &{G-7B} &  \\ 
& {10B}  &{G-10B} &  \\ 
& {20B}  &{G-20B} &  \\ \hline 
\multirow{6}{*}{LLaMA} 
& {1.5B}  &{L-1.5B} &   \\ 
& {2B}  &{L-2B} &  \\ 
& {6B}  &{L-6B} &  \multirow{2}{*}{\shortstack{WikiText\\Alpaca\\SlimPajama}}  \\ 
& {7B}  &{L-7B} &   \\
& {11B}  &{L-11B} &   \\
& {23B}  &{L-23B} &   \\ \hline
\multirow{1}{*}{LLaMA2} 
& {7B}  &{L2-7B} & {WikiText / Alpaca}   \\ \hline
\end{tabular}}
\caption{Models and datasets are used in our experiments.}
\label{model abb}
\end{table}

\subsubsection{Cost Evaluation}  
we conducted the real cost of training 2B and 7B model. By combining the real single-round training speeds of these models and B-F multiplier (defined in the next subsection), we provide an estimated theoretical training cost for GPT2 and LLaMA models on WebText2, containing G-1.6B/5.4B/10B/20B and L-1.5B/6B/11B/23B.

In our experiments, to ensure clarity and consistency, we divided the model into three submodels for all BCD  experiments. This implies that during each iteration, we only update one-third of the full parameter set. Additional experiments detailing various BCD settings are presented in Appendix for further reference.

\subsubsection{Algorithms Setting}
The models include GPT2 and LLaMA series models, shown in Table \ref{model abb}. SGD is configured with a learning rate of 0.1, momentum of 0.9, and weight decay of 1e-5. Adam is used with different settings depending on the model scale, and more detailed configurations are shown in Appendix.

\subsubsection{Multi-GPU Training Setting}  When calculating convergence time, to avoid improper parallel settings, we control the number of cards used to the minimum necessary for running the training tasks (i.e., if fewer computational resources were used, it would not be possible to conduct model training). Specifically, for 7B models, full-parameter training requires two A800 GPUs, while BCD completes training with only one. Resource configurations for smaller models are detailed in Appendix. This setup ensures that the models are trained efficiently without underutilizing or overextending the computational resources available, providing a practical approach to comparing convergence times and training costs.

\subsubsection{Offloading Setting} 
In our experiments, the offloading configuration was set to Stage 1, where the optimizer states were offloaded to the CPU with the \texttt{device} parameter set to \texttt{cpu} and \texttt{pin\_memory} enabled. Additionally, overlap between communication and computation was enabled, along with contiguous gradient storage. 

Since the 0.15B model can fit entirely within the GPU memory, it does not require model offloading. For the 2B model, a dual-GPU offload configuration is both the fastest and the most cost-effective. In contrast, a three-GPU setup provides sufficient memory to fully store the model without the need for offloading. Therefore, in our experiments, we utilized the offload configuration under the dual-GPU mode for the 2B model. This approach optimizes both performance and cost efficiency by balancing the use of computational resources and avoiding unnecessary complexity in configurations where full model storage in GPU memory is feasible.

\begin{table*}
\centering
\renewcommand{\arraystretch}{1.2}
\setlength{\tabcolsep}{6.0pt}
\begin{tabular}{c|ccc|ccc|ccc|ccc}
\hline
\multirow{2}{*}{\textbf{Model}} & \multicolumn{3}{c|}{\textbf{SGD}} & \multicolumn{3}{c|}{\textbf{BCD-SGD}} & \multicolumn{3}{c|}{\textbf{Adam}} & \multicolumn{3}{c}{\textbf{BCD-Adam}} \\ \cline{2-13}
& \textbf{Iter} & \textbf{PPL} & \textbf{Loss}
& \textbf{Iter} & \textbf{PPL} & \textbf{Loss}
& \textbf{Iter} & \textbf{PPL} & \textbf{Loss}
& \textbf{Iter} & \textbf{PPL} & \textbf{Loss} \\ \hline
GPT2-wiki &111587 &6.96 &1.94 &204575 &\textbf{6.91} &\textbf{1.93} &83690 &6.59 &1.88  &232471 &6.62 &1.89\\
GPT2-web  &1045000 &28.73 &3.35 &2770000 &31.28 &3.44 &499000 &28.73 &3.35 &1045000 &30.71 &3.42\\ \hline
\end{tabular}
\caption{Experimental results of the BCD method and full-parameter training with SGD and Adam.
}
\label{Validity Results}
\end{table*}

\begin{table}[!htbp]
\centering
\renewcommand{\arraystretch}{1.2}
\setlength{\tabcolsep}{6.0pt}
\scalebox{0.78}{
\begin{tabular}{c|c|ccc|ccc}
\hline
\multirow{2}{*}{\textbf{Model}} & \multirow{2}{*}{\textbf{Dataset}} 
& \multicolumn{3}{c|}{\textbf{Adam}} & \multicolumn{3}{c}{\textbf{BCD-Adam}} \\ \cline{3-8}
& & \textbf{Iter} & \textbf{PPL} & \textbf{Loss} & \textbf{Iter} & \textbf{PPL} & \textbf{Loss} \\ \hline
G-2B &wiki  & 9968 & 96.69 & 0.70 & 28870 & 97.96 & \textbf{0.73} \\
G-2B &alpaca& 9652 & 28.30 & 0.78 & 25898 & \textbf{27.85} & 0.80 \\
G-2B &slimpajama & 9550 & 69.28 & 3.87 & 28707 & \textbf{58.82} & \textbf{3.59} \\
L-2B &alpaca & 18579 & 21.06 & 0.17 & 20076 & \textbf{21.04} & 0.22 \\
G-7B &alpaca & 9505 &33.19  & 0.78 & 17740 & \textbf{30.55} & 0.81 \\
L-7B &wiki & 8285 & 66.84 & 0.16 & 17879 & \textbf{65.68} & 0.19 \\\hline
\end{tabular}}
\caption{Comparison of the BCD method and full-parameter training across different datasets using the Adam optimizer.}
\label{bcd_adam_results}
\end{table}

\begin{figure}[!ht]
    \centering
    \includegraphics[width=0.5\textwidth]{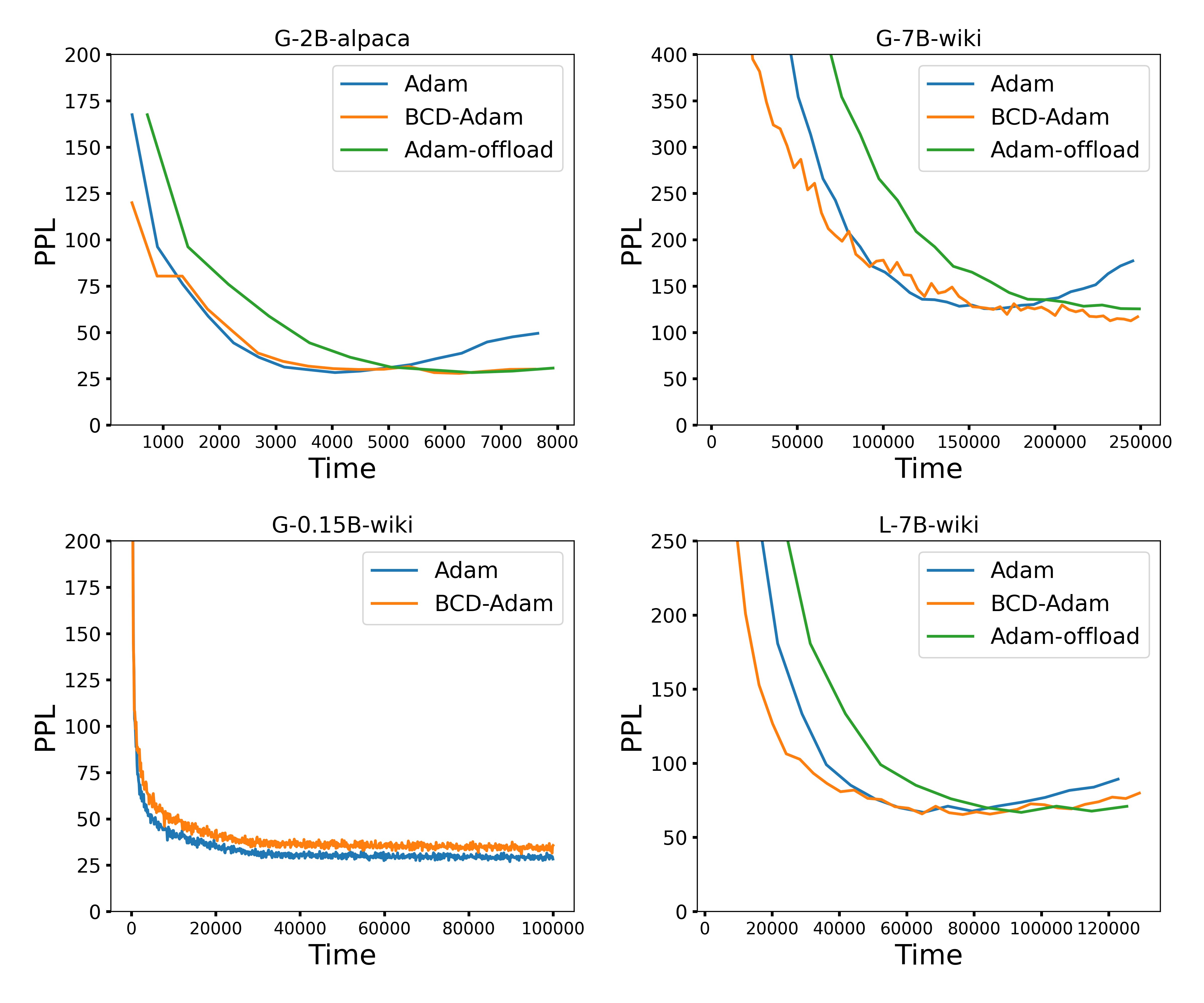}
    \caption{Comparative analysis of perplexity trends across different training methods}
    \label{PPL Trend}
\end{figure}

\begin{figure}[!ht]
    \centering
    \includegraphics[width=0.43\textwidth]{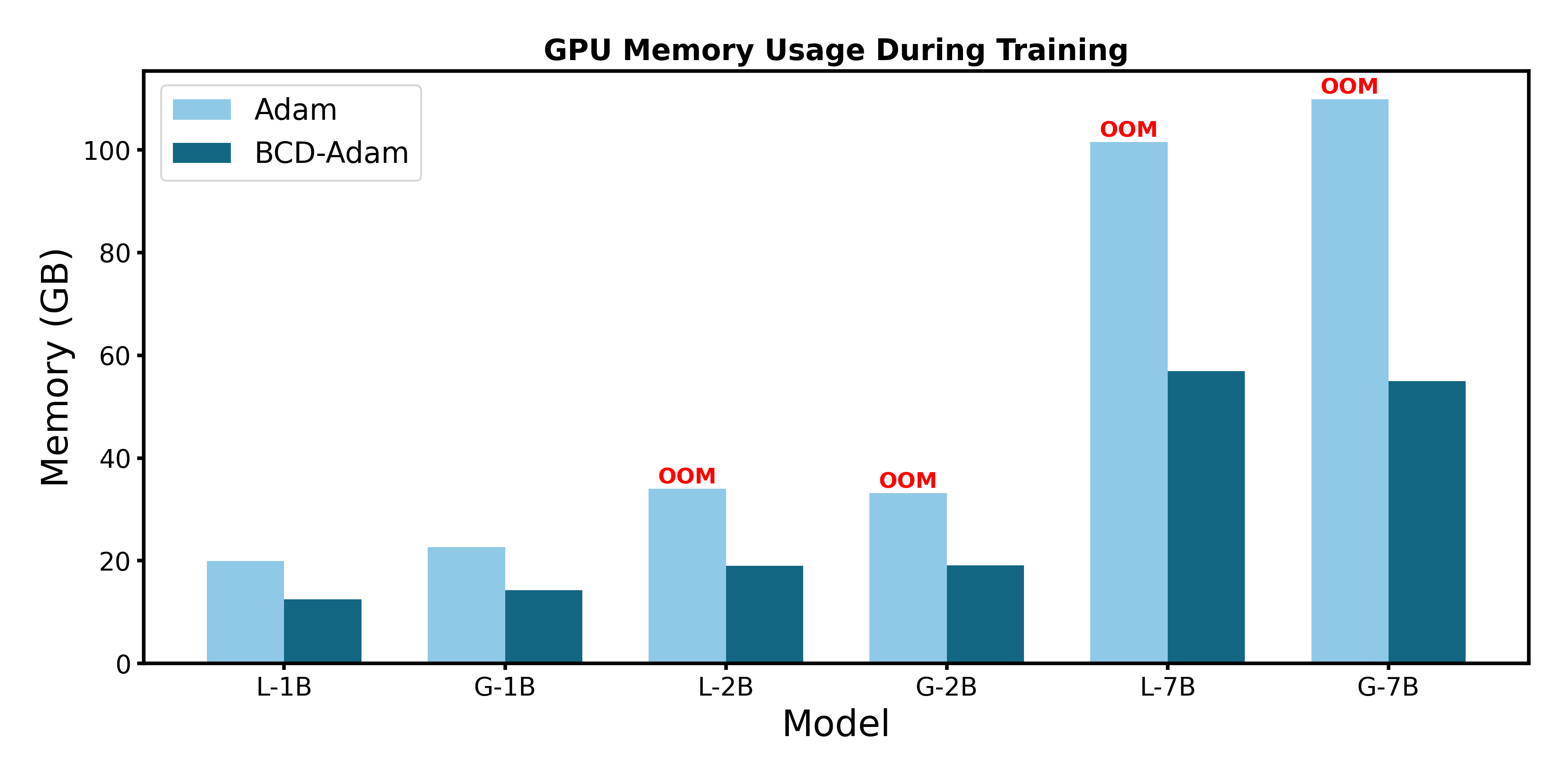}
    \caption{GPU memory usage during training across different models and optimizers.}
    \label{Train memory}
\end{figure}

\subsubsection{Fine-tuning Setting} 
 All models were fine-tuned with a batch size of 16. The learning rate was set to  1e-5 for both LoRA and GaLore. For LoRA and GaLore, a rank of 16 was used. All methods adopted a cosine learning rate scheduler and a weight decay of 1e-5. Additionally, the galore\_scale parameter was set to 2, and lora\_alpha was set to 32.
 
\subsection{{Model Training Analysis}}
This section presents a systematic evaluation of the effectiveness of BCD in training neural networks, with a particular emphasis on its applicability to LLMs. In addition, we conduct a comprehensive analysis of BCD's computational efficiency in large-scale training scenarios. To demonstrate the generalizability of BCD across diverse neural network architectures, we employ it to train different models of varying scales, including ResNet, GPT2, and Llama. Specifically, we present a detailed examination of computational resource utilization during the training of a 7-billion parameter LLM utilizing large-scale computing infrastructure, thereby highlighting the scalability and resource optimization benefits offered by the BCD approach, comparing with distributed Adam and offloading.

\subsubsection{Experimental Results}

In small model experiment, we employed two optimizers, Adam and SGD, and trained the models using both the BCD method and full-parameter updating method. The results are shown in  Table \ref{Validity Results}. Overall, we observe that BCD-based training demonstrates certain advantages. From the perspective of the loss function, the performance of models trained with BCD is superior to those trained with full-parameter updating in most cases. Notably, the loss differences across different models and datasets remain within a narrow margin of 0.1, with BCD occasionally achieving lower loss than full-parameter training. This indicates that BCD can match or even exceed full-parameter methods in optimization efficiency while reducing parameter updates. These results are further validated by additional experiments on ResNet variants, as detailed in Appendix.
 
Table~\ref{bcd_adam_results} presents the results of large-scale model experiments, demonstrating that BCD achieves PPL performance comparable to or even better than full-parameter training in most cases. This suggests strong transferability across models of different scales.

Figure~\ref{PPL Trend} illustrates the trend of Perplexity (PPL) over training time across different models and datasets under various optimization strategies, including Adam, BCD-Adam, and Adam-offload. Across multiple settings, BCD-Adam consistently achieves comparable or lower PPL than full-parameter Adam, particularly on larger models such as G-7B-wiki and L-7B-wiki. Furthermore, the convergence curves of BCD-Adam are noticeably smoother, suggesting improved training stability. These 
results also demonstrate the effectiveness and robustness of BCD across diverse model scales and data domains.

In summary, BCD exhibits training dynamics that closely align with those of full-parameter optimization across a range of model sizes and datasets, while demonstrating superior convergence behavior in several challenging scenarios. These results further substantiate the feasibility and effectiveness of BCD as a scalable, efficient, and principled training paradigm.

\subsubsection{Memory Analysis}
In Figure~\ref{Train memory}, we present the GPU memory usage during training across different model scales using either full-parameter training or our proposed BCD method. For the 1B and 2B models, we used an RTX 4090 GPU, while the 7B models were trained on an A800 GPU (80GB).

As shown in the Figure, the BCD method consistently reduces memory consumption during training across all model sizes. Compared to full-parameter training, it achieves a peak memory reduction of more than 45\%. Bars labeled in red with \texttt{OOM} indicate configurations that exceed the available memory of the hardware and thus cannot complete training due to out-of-memory errors. For instance, the L-2B and G-2B models exceed the 24GB limit of the RTX 4090 under full-parameter training, and the L-7B and G-7B models similarly fail to run on the A800.

In contrast, BCD successfully trains all model configurations without encountering any \texttt{OOM} issues, demonstrating its capability to enable large-scale model training under constrained GPU resources. These results underscore the scalability and memory efficiency of BCD as a practical training strategy for large-scale language models.

\subsubsection{B-F multiplier}
Based on the experimental results, to gain the same performance modes, in Adam experiments, we found that the average number of training epochs required by the BCD method is {1.39} times that of full parameters Adam methods, with the worst-case scenario being  {2.89} times. We set the Adam's average B-F (BCD-Full multiplier)  multiplier  as 1.39 and Adam's the worst  B-F multiplier as 2.89.

\subsection{Economic Cost Analysis}
In this section, we conduct experiments on GPT-2 and LLaMA models with varying parameter scales. Through the analysis of both actual and theoretical training costs, we further demonstrate that the BCD method offers significant economic advantages over traditional training approaches.

\subsubsection{The Rental Costs of GPUs } 
Currently, the market price ratio of 4090, A800, and A100 is approximately 1:2.5:4, with more detailed pricing information provided in Appendix. In this section, we refer to the rental costs from Infinigence AI, Ltd., which are as follows: \$0.29/hour for a single RTX 4090 GPU, \$1.20/hour for a single A100 GPU, and \$0.69/hour for a single A800 GPU. The total training cost is the product of the number of training rounds, the training time per round, the number of GPUs used, and the rental cost. The GPU hour details on different platform can be found in Appendix.

\subsubsection{Training Cost} 
Table~\ref{real cost} compares BCD and full-parameter training in terms of GPU count, training time (in hours), and monetary cost (in USD) across RTX 4090, A800, and A100 platforms. Overall, while the BCD method typically incurs longer training durations due to its block-wise update mechanism, it achieves substantial cost reductions in most scenarios. For instance, when training the L-7B model on the Wiki dataset, BCD reduces the cost from \$43.74 to \$31.22 on A100, and from \$26.30 to \$18.74 on A800. Even on consumer-grade GPUs such as the RTX 4090, it slightly lowers the cost from \$3.37 to \$3.29, demonstrating that BCD provides cost benefits on both high-performance and consumer-grade hardware. Results on the G-7B model further support this trend. On the Wiki dataset, BCD reduces the cost from \$70.08 to \$41.36, despite a moderate increase in training time from 50.66 to 59.79 hours. On the Alpaca dataset, both G-7B and L-7B benefit from consistent cost savings, with reductions from \$18.49 to \$12.21 and from \$26.39 to \$17.63, respectively. In summary, BCD provides a cost-effective solution without sacrificing model performance. This makes it well-suited not only for resource-limited environments but also for cost-sensitive training tasks where time efficiency is less critical.
\begin{table}[!htbp]
\centering
\renewcommand{\arraystretch}{1.2}
\setlength{\tabcolsep}{6.0pt}
\scalebox{0.67}{
\begin{tabular}{c|c|ccc|ccc|c}
\hline
\multirow{2}{*}{\textbf{Model}} & \multirow{2}{*}{\textbf{Dataset}} 
& \multicolumn{3}{c|}{\textbf{Adam}} 
& \multicolumn{3}{c|}{\textbf{BCD-Adam}} 
& \multirow{2}{*}{\textbf{Platform}} \\ \cline{3-8}
& & \textbf{GPU} & \textbf{Cost} & \textbf{Time} 
& \textbf{GPU} & \textbf{Cost} & \textbf{Time} & \\ \hline
G-2B &slimpajama &3 &6.08 &6.99 &2 &7.30 &12.57 &4090 \\ 
G-2B &slimpajama &1 &49.59 &40.95 &1 &116.53 &96.22 &A100\\ 
L-7B &wiki       &2 &26.30 &19.01 &1 &\textbf{18.74} &27.10 &A800\\
L-7B &wiki       &8 &3.37 &1.45 &4 &\textbf{3.29} &2.84  &4090 \\ 
L-7B &wiki       &2 &43.74 &18.06 & 1&\textbf{31.22} &25.78 &A100\\
G-7B &wiki   &2 &70.08 &50.66  &1 &\textbf{41.36} &59.79  &A800\\ 
G-7B &alpaca &2  &18.49 &13.37 &1 &\textbf{12.21} &17.66 &A800\\ 
L-7B &alpaca &2  &26.39 &19.08 &1 &\textbf{17.63} &25.48 &A800\\ \hline
\end{tabular}}
\caption{
Monetary cost and time comparison of BCD and full-parameter training on different GPUs.
}
\label{real cost}
\end{table}

In addition to empirical evaluations based on actual training runs, we provide a theoretical estimation of total training cost under various hardware configurations to systematically demonstrate the economic advantages of the BCD method. The total training cost is defined as the product of per-iteration training time, number of training iterations, number of GPUs utilized, and hourly rental price per GPU.

Since BCD updates only a subset of parameters in each iteration, it generally requires more training steps than full-parameter training. Accordingly, we incorporate both average-case and worst-case B-F multipliers as defined previously. For instance, in the average-case scenario, full-parameter training of the GPT2-10B model on an A100 cluster (1-node, 4 GPUs), resulting in a per-iteration time of 5772 ms and a total estimated cost of \$11.6.

In comparison, the BCD method employs a node configured with 8 A100 GPUs, achieving a significantly shorter per-iteration time of 2414 ms. The total cost is reduced to approximately \$8.7—amounting to 74.9\% of the full-parameter baseline. Similar analyses are conducted across model sizes from 1.6B to 20B on A100, A800, and RTX 4090 clusters. The relative cost reductions, expressed as percentages, are summarized in Appendix.

\subsection{Comparison with Fine-Tuning}
We conducted a series of experiments to evaluate the effectiveness of the proposed BCD method for fine-tuning large language models. Specifically, we selected L2-7B and G-1.5B as base models and utilized two small-scale training datasets: WikiText and Alpaca. The proposed BCD approach was compared against three mainstream fine-tuning strategies: full fine-tuning (full-parameter updating method in ealier experiment), GaLore, and LoRA. Evaluation was conducted using two standard metrics: training loss and validation loss.

As presented in Table~\ref{compare with finetune}, BCD consistently demonstrates lower training losses across most configurations, with particularly significant improvements on the L2-7B model. On the WikiText and Alpaca datasets, BCD achieves training losses of 0.166 and 0.192 respectively, substantially outperforming full fine-tuning, GaLore, and LoRA methods. These results clearly demonstrate BCD's superior model fitting capability. Under the G-1.5B + WikiText setting, BCD yields a noticeably higher evaluation loss (4.35) compared to LoRA (0.613) and full fine-tuning (4.370), indicating a potential overfitting issue in this specific configuration. However, for the larger L2-7B model, BCD exhibits more stable generalization. Notably, on the Alpaca dataset, it achieves a evaluation loss of 1.192, outperforming both LoRA (1.931) and full fine-tuning (1.269).
Overall, BCD delivers performance comparable to or better than full fine-tuning in most cases, highlighting its practical potential as an efficient and effective fine-tuning strategy for large models.

\begin{table}[!htbp]
\centering
\renewcommand{\arraystretch}{1.2}
\setlength{\tabcolsep}{6.0pt}
\scalebox{0.65}{
\begin{tabular}{c|c|cc|cc|cc|cc}
\hline
\multirow{2}{*}{\textbf{Model}} & \multirow{2}{*}{\textbf{Dataset}} 
& \multicolumn{2}{c|}{\textbf{LoRA}} 
& \multicolumn{2}{c|}{\textbf{GaLore}} 
& \multicolumn{2}{c|}{\textbf{BCD-Adam}} 
& \multicolumn{2}{c}{\textbf{Adam}} \\ \cline{3-10}
& & \textbf{Train} & \textbf{Eval} 
  & \textbf{Train} & \textbf{Eval} 
  & \textbf{Train} & \textbf{Eval} 
  & \textbf{Train} & \textbf{Eval} \\ \hline
G-1.5B &wiki   &0.34  &0.61  &1.79  &2.18  & \textbf{0.22} &4.35  &0.39  &4.37\\
L2-7B  &wiki   &0.86  &1.68  &0.87  &1.37  & \textbf{0.16} &1.89  &0.12  &1.90\\
L2-7B  &alpaca &0.81  &1.93  &0.71  &0.87  & \textbf{0.19} &1.19  &0.12  &1.26\\\hline
\end{tabular}}
\caption{
Training and evaluation losses of LoRA, GaLore, BCD, and full-parameter update on WikiText and Alpaca with different model sizes.
}
\label{compare with finetune}
\end{table}
\subsection{Ablation Experiments}
This section investigates the impact of submodel scale per training iteration on the final convergence behavior of the BCD algorithm. Specifically, we conduct experiments by unfreezing varying proportions of model parameters, which we  referred to as the Unfrozen Parameter Ratio (abbr. UFP), and systematically evaluate the resulting convergence performance and memory footprint.

\begin{table}[!ht]
\centering
\renewcommand{\arraystretch}{1.2}
\setlength{\tabcolsep}{6.0pt}
\scalebox{0.7}{
\begin{tabular}{c|cccccc}
\hline
{\textbf{\#UFP}} &{\textbf{Opt}}  &{\textbf{\#epoch}} &{\textbf{PPL}} &{\textbf{Loss}}      &{\textbf{Mem}} \\ \hline
{1}     &{SGD}  &{24}  &{6.96}  &{1.94}    &{1824.45MB}     \\ 
{1/2}   &{SGD}  &{40}  &{6.71}  &{1.90}    &{1216.30MB}      \\ 
{1/3}   &{SGD}  &{44}  &{6.91}  &{1.93}    &{1013.58MB}      \\ 
{1/4}   &{SGD}  &{50}  &{6.75}  &{1.91}    &{912.23MB}      \\ 
{1}     &{Adam} &{18}  &{6.59}  &{1.88}    &{2432.60MB}      \\ 
{1/2}   &{Adam} &{40}  &{6.63}  &{1.89}    &{1520.38MB}      \\ 
{1/3}   &{Adam} &{50}  &{6.62}  &{1.89}    &{1216.30MB}     \\ 
{1/4}   &{Adam} &{60}  &{6.68}  &{1.89}    &{1064.26MB}      \\ \hline
\end{tabular}}
\caption{The performance of G-0.15B  models with different scales of frozen parameters under various optimizers.}
\label{UFP}
\end{table}

Table~\ref{UFP}  shows that reducing the number of UFP significantly decreases memory consumption. Overall, a clear trade-off is observed between memory efficiency and training efficiency: reducing the UFP significantly lowers peak memory usage, but generally requires more epochs to converge. For instance, under SGD optimization, decreasing UFP from 1 to 1/4 reduces memory consumption by approximately 50\% (from 1824.45 MB to 912.23 MB), at the cost of nearly doubling the number of epochs needed for convergence (from 24 to 50 epochs).

In terms of final model performance, measured by PPL and Loss, moderate UFP values tend to yield the best results. With SGD, the lowest PPL (6.71) is achieved at UFP=1/2. Under Adam, the minimal PPL (6.62) occurs at UFP=1/3, suggesting that finer-grained parameter updates may better leverage Adam’s adaptive learning capabilities. Notably, while full-parameter updates (UFP=1) with Adam achieve competitive PPL (6.59), they incur substantially higher memory overhead (2432.60 MB), which may be prohibitive in resource-constrained settings. However, these performance fluctuations are extremely minor and may be overshadowed by random noise, having negligible impact in practical engineering scenarios.


In summary, the choice of UFP primarily affects the convergence speed of the model, while having minimal impact on the final convergence outcome.  In terms of iteration dynamics, smaller UFP values require more training epochs, but each epoch is faster and consumes less memory. Therefore, the selection of UFP should be carefully balanced according to specific requirements on training time or hardware resources. Overall, despite differences in training efficiency, the influence of UFP on final model performance is limited, demonstrating that the BCD algorithm exhibits strong robustness and stability across different parameter update ratios.
\section{Conclusion}
In this paper, we present a BCD framework, enhanced with engineering optimizations, to enable cost-efficient training of large-scale models on clusters characterized by high performance, high communication overhead, and low economic cost. This framework is particularly promising when applied to retired data-center GPUs (e.g., V100), consumer-grade GPUs, or cost-efficient AI clusters such as the Hygon
DCU, thereby lowering the hardware barrier for large-scale model development.



\bibliography{aaai2026}
\appendix
\section{Appendix}
\section{Experiments}
We conduct a series of experiments to evaluate the proposed BCD framework from multiple perspectives, including training efficiency, memory usage, cost-effectiveness, and model performance. Specifically, we report results on parameter freezing, dataset compression, full training analysis, GPU hour savings, and comparisons with offloading and distributed training baselines.

\section{Model configurations}
Our experiments involve two categories of language models, GPT-2 and LLaMA, with four variants from each: G-1.6B, G-5.4B, G-10B, G-20B, and L-1.5B, L-6.0B, L-11B, L-23B. The parameter scales, architectural configurations, and memory usage of these models are summarized in Table~\ref{modelset}. The variable \textbf{L} denotes the number of Transformer layers, \textbf{H} represents the hidden size, and \textbf{A} indicates the number of attention heads. \textbf{Mem} refers to the peak memory usage during training. Notably, the L-1.5B model employs a feed-forward network (FFN) with a size of 5,050, in contrast to other LLaMA models, which utilize an FFN size of 9,200. Table~\ref{Distribution gpu} shows the Multi-GPU training setup section by detailing the resource configurations employed for training smaller models.

To ensure fairness and comparability, we maintained consistent hyperparameter settings across all experimental setups: the learning rate of 5e-5, the batch size of 64, and the Adam optimizer with momentum coefficients m = 0.9 and v = 0.95. All experiments were conducted using FP32 precision to eliminate variability introduced by mixed- or low-precision computation, thereby ensuring reliable evaluation. 

\begin{table}[!htbp]
\centering
\renewcommand{\arraystretch}{1.2}
\setlength{\tabcolsep}{6.0pt}
\scalebox{0.9}{
\begin{tabular}{c|cccccc}
\hline
\textbf{Model} & \textbf{GPU} & \textbf{L} & \textbf{H} & \textbf{A} & \textbf{Mem} \\ \hline
{G-1.6B}  &{1}  &{30} &{2048} &{16} &{24.077 GB}     \\ 
{G-5.4B}  &{4}  &{28} &{3968} &{32} &{81.881 GB}     \\ 
{G-10B}  &{8}  &{56} &{3968} &{32} &{160.738 GB}     \\ 
{G-20B}  &{16}  &{110} &{3968} &{32} &{312.805 GB}     \\ 
{L-1.5B} &{1}   &{30} &{2048} &{16} &{24.632 GB}   \\ 
{L-6.0B} &{4}  &{32} &{4096} &{32} &{89.815 GB}   \\
{L-11B} &{8}  &{64} &{4096} &{32} &{175.714 GB}    \\ 
{L-23B} &{16} &{128} &{4096} &{32} &{347.539 GB}      \\ \hline
\end{tabular}}
\caption{Model configurations and memory usage.}
\label{modelset}
\end{table}

\begin{table}[!htbp]
\centering
\renewcommand{\arraystretch}{1.2}
\setlength{\tabcolsep}{6.0pt}
\scalebox{0.9}{
\begin{tabular}{c|cccccc}
\hline
{\textbf{Update Type}} & {\textbf{Model}} & {\textbf{GPU}} & {\textbf{GPU Type}}  \\ \hline
\multirow{6}{*}{Full-parameter} 
& {G-0.16B} & {1} & {RTX 4090} \\
& {L-2B} & {1} & {A800} \\
& {G-1.6B} & {2} & {RTX 4090} \\
& {G-2B} & {3} & {RTX 4090} \\
& {G-7B} & {2} & {A800} \\ 
& {L-7B} & {2} & {A800} \\  \hline
\multirow{6}{*}{BCD} 
& {G-0.16B} & {1} & {RTX 4090} \\
& {L-2B} & {1} & {A800} \\
& {G-1.6B} & {1} & {RTX 4090} \\
& {G-2B} & {2} & {RTX 4090} \\
& {G-7B} & {1} & {A800} \\
& {L-7B} & {1} & {A800} \\   \hline
\end{tabular}}
\caption{Settings for BCD and full-parameter training.}
\label{Distribution gpu}
\end{table}

\section{UFP experiments}

In this experiment, UFP (Unfrozen Parts) denotes the proportion of model parameters that remain trainable during training. By freezing a certain proportion of parameters (e.g., 1/2, 1/3, 1/4, 1/5), we examine how parameter freezing influences memory usage and training efficiency. Freezing a subset of parameters reduces both gradient computation and optimizer state storage, which in turn lowers memory consumption. However, this strategy may also impact training efficiency and degrade final model performance. This section investigates the memory and time efficiency of GPT-2 and LLaMA models under different UFP configurations.

\begin{figure}[!ht]
    \centering
    \includegraphics[width=0.45\textwidth]{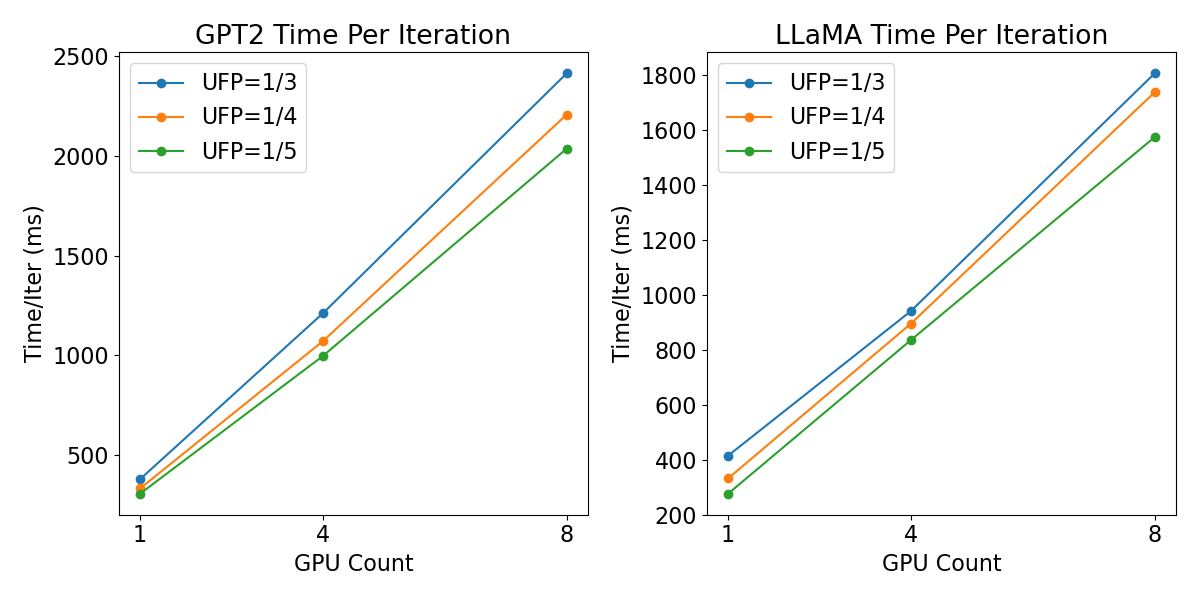}
    \caption{Average training time per iteration (ms) for GPT-2 and LLaMA models under different UFP settings using the BCD method.}
    \label{tabelx2}
\end{figure}

\begin{table}[!htbp]
\centering
\renewcommand{\arraystretch}{1.2}
\setlength{\tabcolsep}{6.0pt}
\scalebox{0.95}{
\begin{tabular}{c|ccccc}
\hline
\textbf{Model} & \textbf{GPU} & \textbf{T-1/3} & \textbf{T-1/4} & \textbf{T-1/5} \\ \hline
G-1.6B  & 1      & 0.37  & 0.33  & 0.30 \\ 
G-5.4B  & 4      & 1.21  & 1.07  & 0.99 \\ 
G-10B   & 8      & 2.41  & 2.20  & 2.03 \\
L-1.5B  & 1      & 0.41  & 0.33  & 0.27 \\
L-6.0B  & 4      & 0.94  & 0.89  & 0.83 \\ 
L-11B   & 8      & 1.80  & 1.73  & 1.57 \\ \hline

\end{tabular}}
\caption{Average per-iteration training time (s) for GPT-2 and LLaMA models under different UFP settings using the BCD method. T-1/3 refers to the average time when one-third of the parameters are unfrozen.}
\label{UFP-345}
\end{table}

\subsection{Analysis}
In Figure~\ref{tabelx2}, the experimental results demonstrate that increasing the proportion of frozen parameters leads to a consistent decrease in the time per iteration. This trend is observed across various model scales and GPU configurations, confirming its generality.

The results in Table~\ref{UFP-345} demonstrate that, for both GPT-2 and LLaMA models, freezing a larger portion of parameters reduces the computation time per iteration. This effect is more pronounced in smaller models. These findings indicate that, in resource-constrained settings, parameter freezing can effectively reduce memory consumption and per-iteration computational cost. While this approach may necessitate additional iterations to achieve convergence, the overall computational savings remain notable.

This highlights that, for large-scale model training under limited resources, adopting a suitable parameter-freezing strategy can substantially alleviate per-iteration computational burden. Still, achieving an optimal balance between total training time and final model performance requires careful tuning of the freezing ratio.

\section{Extended evaluation on ResNet models}

To further validate the correctness and generality of the BCD method beyond language models, we conduct additional experiments on a series of ResNet models using both SGD and Adam optimizers. As shown in Table~\ref{Validity Results 2}, the final accuracy and loss values of BCD-trained models are comparable to, or even slightly better than, those obtained from full-parameter training. These results demonstrate that the BCD method maintains training stability and effectiveness across different network architectures and optimization algorithms.

\begin{table*}[!htbp]
\centering
\renewcommand{\arraystretch}{1.2}
\setlength{\tabcolsep}{6.0pt}
\scalebox{1.0}{
\begin{tabular}{c|ccc|ccc|ccc|ccc}
\hline
\multirow{2}{*}{\textbf{Model}} & \multicolumn{3}{c|}{\textbf{SGD}} & \multicolumn{3}{c|}{\textbf{BCD-SGD}} & \multicolumn{3}{c|}{\textbf{Adam}} & \multicolumn{3}{c}{\textbf{BCD-Adam}} \\ \cline{2-13} 
& \textbf{Epoch} & \textbf{Acc} & \textbf{loss} & \textbf{Epoch} & \textbf{Acc} & \textbf{loss}& \textbf{Epoch} & \textbf{Acc} & \textbf{loss}& \textbf{Epoch} & \textbf{Acc} & \textbf{loss}\\ \hline
ResNet-8 &639 &86.9 &0.1671  &\textbf{395} &\textbf{87.2} &0.1783  &389 &83.1 &0.3955   &\textbf{309} &\textbf{86.0} &\textbf{0.2279} \\
ResNet-14 &323 &88.9 &0.0708 &391 &\textbf{89.0} &\textbf{0.0613}  &491 &85.7 &0.0969   &\textbf{372} &\textbf{87.6} &\textbf{0.0527}\\
ResNet-20 &191 &88.7 &0.0574  &322 &88.7 &\textbf{0.0263} &768 &86.3 &0.0269   &\textbf{423} &\textbf{88.3} &\textbf{0.0222} \\
ResNet-50 &273 &90.4 &0.0260  &\textbf{251} &88.9 &\textbf{0.0120} &/&/&/&/&/&/\\
ResNet-101 &286 &70.9 &0.0691 &376 &70.1 &\textbf{0.0411} &/&/&/&/&/&/\\ \hline
\end{tabular}}
\caption{
Experimental results of the BCD method and full-parameter training with SGD and Adam.
}
\label{Validity Results 2}
\end{table*}

\section{Dataset compress experiments}

This section investigates the effect of reducing the training dataset size on the per-epoch training scale in BCD. According to the Scaling Law, optimal model performance requires a proportional relationship among computational resources, dataset size, and model size. Based on this principle, when training each submodel, the amount of training data should be scaled down in accordance with the submodel’s size. This experiment is designed to empirically test this hypothesis.

\subsubsection{Reducing training dataset}
According to the scaling law, reducing model size allows for a proportional reduction in the amount of training data needed to maintain a balance between computational efficiency and model performance. Within the BCD framework, training can be viewed as operating on a smaller submodel, which, by the scaling law, does not require the full dataset originally used for the full model. Consequently, scaling down the training data in line with the submodel size becomes a feasible strategy for reducing inference overhead and improving computational efficiency.


To fully exploit the available data while reducing training costs, the BCD method adopts a sampling-based strategy. For each submodel, a distinct subset of the original training dataset is sampled. As training progresses across different submodels, non-overlapping subsets are employed, ensuring broad data coverage over time. This design enables comprehensive data utilization throughout the training process while substantially reducing the computational burden associated with large-scale datasets.

\subsubsection{Dataset setting}
We use the CIFAR-10 and WikiText datasets as benchmarks. In each epoch, 90\% of the training data is re-sampled from the full training set. The results obtained using the full training dataset are also reported as baselines for comparison.

\begin{table}[!htbp]
\centering
\renewcommand{\arraystretch}{1.2}
\setlength{\tabcolsep}{5.0pt}
\scalebox{0.9}{
\begin{tabular}{c|cccccc} \hline
\textbf{SR} & \textbf{UFP} & \textbf{Opt} & \textbf{Epoch} & \textbf{PPL} & \textbf{Loss} & \textbf{Mem} \\ \hline
{0.9}  &{1/3}    &{SGD} &{40}  &{6.89}  &{1.9312}    &{1.01GB}     \\ 
{1}    &{1/3}    &{SGD} &{44}  &{6.91}  &{1.9343}    &{1.01GB}     \\ 
{1}    &{1}      &{SGD} &{24}  &{6.96}  &{1.9407}    &{1.82GB}     \\ 
{0.9}  &{1/3}    &{Adam}&{60}  &{6.75}  &{1.9110}    &{1.21GB}    \\ 
{1}    &{1/3}    &{Adam}&{50}  &{6.62}  &{1.8910}    &{1.21GB}   \\ 
{1}    &{1}      &{Adam}&{18}  &{6.59}  &{1.8869}    &{2.43GB}     \\ \hline
\end{tabular}}
\caption{The GPT-2 model's performance with different training dataset sizes and training methods.}
\label{GPT2 90data}
\end{table}
 
\begin{table}[!htbp]
\centering
\renewcommand{\arraystretch}{1.2}
\setlength{\tabcolsep}{6.0pt}
\scalebox{1.0}{
\begin{tabular}{c|cccccc}
\hline
\textbf{SR} & \textbf{UFP} & \textbf{Epoch} & \textbf{Acc} & \textbf{Loss} & \textbf{Mem} \\ \hline
{0.9}  &{1/3}     &{427}    &{88.7\%}  &{0.0616} &{1.12MB}     \\ 
{1}    &{1/3}     &{391}    &{89.0\%}  &{0.0613} &{1.12MB}     \\ 
{1}    &{1}       &{323}    &{88.9\%}  &{0.0708} &{2.01MB}     \\\hline
\end{tabular}}
\caption{The ResNet-14 model performance with different training dataset sizes and different training methods.}
\label{ResNet14 90data}
\end{table}

\subsubsection{Analysis}
As shown in Table~\ref{GPT2 90data} and Table~\ref{ResNet14 90data}, reducing the training dataset by 10\% has minimal impact on model performance under the BCD framework. Compared to full-data training, using 90\% of the dataset leads to only marginal accuracy degradation. For instance, in GPT-2 experiments with the SGD optimizer, the model trained on the full dataset achieves 69.1\% accuracy, while the model trained on 90\% of the data reaches 68.9\%, indicating a negligible 0.2\% drop. Notably, with the Adam optimizer, the 90\%-sampled model slightly outperforms the full-data model (67.5\% vs. 66.2\%), suggesting that moderate data reduction tends to enhance optimizer stability during convergence in certain cases.

For ResNet-14, a 10\% dataset reduction yields only a 0.3\% accuracy decline (from 89.0\% to 88.7\%), while the loss remains virtually unchanged (0.0613 vs. 0.0616), indicating a similarly negligible effect on convolutional architectures. Moreover, models trained on the full dataset typically converge in fewer epochs, suggesting that data reduction may slightly delay convergence but has minimal impact on final model performance.

\subsection{GPU hour analysis}

\begin{table}[!htbp]
\centering
\renewcommand{\arraystretch}{1.2}
\setlength{\tabcolsep}{6.0pt}
\scalebox{0.9}{
\begin{tabular}{c|ccccccccc}
\hline
\multirow{2}{*}{ } & \multicolumn{1}{c|}{\textbf{G-1.6B}} & \multicolumn{1}{c|}{\textbf{G-5.4B}} & \multicolumn{1}{c|}{\textbf{G-10B}} & \multicolumn{1}{c}{\textbf{G-20B}} \\ \cline{2-5} 
&\multicolumn{1}{c|}{\textbf{GPUh }}  &\multicolumn{1}{c|}{\textbf{GPUh }}
&\multicolumn{1}{c|}{\textbf{GPUh }}
&\multicolumn{1}{c}{\textbf{GPUh }} \\ \hline

 \scalebox{1}{Full Update}  &{--}   &{--}  &{--}   &{--}  \\ 
\scalebox{1}{BCD(The worst)}  &-56.1\%   &-13.3\%   &+2.8\%   &-29.4\%  \\ 
\scalebox{1}{BCD(Average)}  &-80.2\%   &-60.9\%  &-53.6\% &-68.2\% \\ \hline
\end{tabular}}
\caption{Reduction in GPU hours using BCD compared to full parameter updates across different models on an RTX 4090 cluster.}
\label{gpuh1}
\end{table}

\begin{table}[!htbp]
\centering
\renewcommand{\arraystretch}{1.2}
\setlength{\tabcolsep}{6.0pt}
\scalebox{0.90}{
\begin{tabular}{c|ccccccccc}
\hline
\multirow{2}{*}{\textbf{}} & \multicolumn{1}{c|}{\textbf{G-1.6B}} & \multicolumn{1}{c|}{\textbf{G-5.4B}} & \multicolumn{1}{c|}{\textbf{G-10B}} & \multicolumn{1}{c}{\textbf{G-20B}} \\ \cline{2-5} 
&\multicolumn{1}{c|}{\textbf{GPUh }}  &\multicolumn{1}{c|}{\textbf{GPUh }}
&\multicolumn{1}{c|}{\textbf{GPUh }}
&\multicolumn{1}{c}{\textbf{GPUh }} \\ \hline
 \scalebox{1}{Full Update}  &{--}   &{--}  &{--}  &{--} \\ 
\scalebox{1}{BCD(The worst)}  &-1.5\%   &+127.0\%   &+131.8\%   &+78.0\%  \\ 
\scalebox{1}{BCD(Average)}  &-55.6\%   &+2.4\%  &+4.5\% &-19.7\% \\ \hline
\end{tabular}}
\caption{Reduction in GPU hours using BCD compared to full parameter updates across different models on an A100 cluster.}
\label{gpuh2}
\end{table}

\begin{table}[!htbp]
\centering
\renewcommand{\arraystretch}{1.2}
\setlength{\tabcolsep}{6.0pt}
\scalebox{0.9}{
\begin{tabular}{c|ccccccccc}
\hline
\multirow{2}{*}{\textbf{}} & \multicolumn{1}{c|}{\textbf{G-1.6B}} & \multicolumn{1}{c|}{\textbf{G-5.4B}} & \multicolumn{1}{c|}{\textbf{G-10B}} & \multicolumn{1}{c}{\textbf{G-20B}} \\ \cline{2-5} 
&\multicolumn{1}{c|}{\textbf{GPUh }}  &\multicolumn{1}{c|}{\textbf{GPUh }}
&\multicolumn{1}{c|}{\textbf{GPUh }}
&\multicolumn{1}{c}{\textbf{GPUh }} \\ \hline
 \scalebox{1}{Full Update}  &{--}   &{--}  &{--}  &{--} \\ \hline
\scalebox{1}{BCD(The worst)}  &-6.4\%   &+114.7\%   &+121.6\%   &+70.5\%  \\ \hline
\scalebox{1}{BCD(Average)}  &-57.8\%   &-3.2\%  &-0.08\% &-23.1\% \\ \hline
\end{tabular}}
\caption{Reduction in GPU hours using BCD compared to full parameter updates across different models on an A800 cluster.}
\label{gpuh3}
\end{table}

Based on the results reported in Tables \ref{gpuh1}, \ref{gpuh2}, and \ref{gpuh3}, the performance of the BCD method varies across model scales, exhibiting both notable benefits and certain limitations. For the G-1.6B and G-20B models, BCD substantially reduces GPU hours compared to full-parameter training. In contrast, for the G-5.4B and G-10B models, their GPU computation time is comparable to—or even slightly exceeds—that of full-parameter training. These observations indicate that there is still potential to improve the computational efficiency of the BCD method.

\section{Performance experiments\label{perf exp}}

In this experiment, we compare the performance of the BCD method, the offloading, and the distributed training approaches within a single iteration. Additionally, we report the estimated time required to fully train the models using these methods, based on the B-F multiplier. In this section, we use the Adam optimizer exclusively, as it is the mainstream choice for training LLMs, and offloading specifically provides an optimized implementation for Adam.

\subsection{Experimental result}

\begin{table}[!htbp]
\centering
\renewcommand{\arraystretch}{1.2}
\setlength{\tabcolsep}{6.0pt}
\scalebox{0.95}{
\begin{tabular}{c|ccccc}
\hline
\textbf{Model} & \textbf{GPU} & \textbf{Offloading} & \textbf{BCD-Adam} \\ \hline

{G-1.6B}  & {1} & 2.40 s & 0.37 s\\
{G-5.4B}  & {4} & 4.37 s & 1.21 s\\ 
{G-10B}  & {8} & 7.07 s& 2.41 s\\ 
{L-1.5B} & {1} & 4.82 s& 0.41 s\\ 
{L-6.0B} & {4} & 6.36 s& 0.94 s\\ 
{L-11B} & {8} & 7.29 s& 1.80 s\\ \hline

\end{tabular}}
\caption{The time comparison between offloading and BCD in one iteration. }
\label{compare offload bcd}
\end{table}

\begin{table}[!htbp]
\centering
\renewcommand{\arraystretch}{1.2}
\setlength{\tabcolsep}{6.0pt}
\scalebox{0.80}{
\begin{tabular}{c|ccc|ccc}
\hline
\multirow{2}{*}{\textbf{Model}} & \multicolumn{3}{c|}{\textbf{Distributed}} & \multicolumn{3}{c}{\textbf{BCD-Adam}} \\ \cline{2-7} 
& \textbf{N/G} & \textbf{Mem} & \textbf{Time} & \textbf{N/G} & \textbf{Mem} & \textbf{Time} \\ \hline
G-1.6B  & 1/2 & 24.077  & 0.57 s & 1/1 & 12.038  & 0.37 s\\
G-5.4B  & 2/4 & 81.881  & 1.93 s & 1/4 & 40.942  & 1.21 s\\
G-10B   & 4/4 & 160.738 & 3.17 s & 1/8 & 80.370  & 2.41 s\\
G-10B   & 2/8 & 160.738 & 3.25 s & 1/8 & 80.370  & 2.41 s\\
G-20B   & 4/8 & 312.805 & 6.37 s & 2/8 & 156.402 & 3.60 s\\
L-1.5B  & 1/2 & 24.632  & 0.42 s & 1/1 & 11.662  & 0.41 s\\
L-6.0B  & 2/4 & 89.815  & 1.28 s & 1/4 & 44.908  & 0.94 s\\
L-8.0B  & 2/6 & 120.327 & 1.65 s & 1/6 & 60.163  & 1.19 s\\
L-11B   & 4/4 & 175.714 & 2.51 s & 1/8 & 87.858  & 1.80 s\\
L-23B   & 4/8 & 347.539 & 4.90 s & 2/8 & 173.769 & 3.47 s \\ \hline

\end{tabular}}
\caption{The time comparison between distributed method and BCD in one iteration. N/G means the number of nodes and the number of GPU per node. Mem is measured in GB.}
\label{compare dist bcd}
\end{table}

\subsubsection{The comparison between offloading and BCD}
Based on the experimental results shown in Table \ref{compare offload bcd}, there is a substantial performance gap between the offloading and BCD methods on GPT-2 and LLaMA models, particularly in terms of time efficiency and scalability. Each method exhibits distinct characteristics, with our approach outperforming offloading in both aspects.

First, the iteration time of the offloading is significantly longer than that of BCD across all GPU configurations. For instance, when using 8 GPUs on GPT-2, the average iteration time of offloading reaches 7.07 s, while BCD requires only 2.41 s—resulting in an approximate 65.8\% reduction. Similarly, for the LLaMA model, offloading’s iteration time increases steeply with the number of GPUs, rising from 4.82 s on a single GPU to 7.29 s on 8 GPUs. In contrast, BCD exhibits a more gradual increase, from 0.41 s to 1.80 s under the same settings. These results demonstrate that Offloading suffers from limited scalability in multi-GPU environments, while BCD adapts more effectively to increasing GPU counts.

The performance bottleneck of offloading primarily stems from frequent data transfers between the CPU and GPU. While offloading reduces GPU memory usage by moving certain model parameters or activation values to the CPU, this strategy incurs significant communication overhead. This issue is further exacerbated in multi-GPU settings, where communication costs escalate with the number of devices involved. Consequently, offloading’s time efficiency deteriorates as the scale of distributed training grows.

In contrast, BCD avoids frequent inter-device communication by freezing a subset of parameters in GPU memory and updating only the necessary ones. This design substantially reduces communication overhead and allocates more GPU resources to core training computations. As a result, BCD demonstrates superior scalability in multi-GPU scenarios.

It is worth noting that even with offloading mode enabled, current implementations are unable to fully utilize GPU resources for pipelined training of models exceeding 10 billion parameters. Although, in theory, Stage 3 offloading could facilitate data parallelism for such large models, this functionality is not yet supported by any official interface. Furthermore, the involvement of cross-node training in these large-scale settings leads to additional performance degradation.

\subsubsection{The comparison between distributed method and BCD}
As shown in Table~\ref{compare dist bcd}, for the G-1.6B model, the iteration time of the BCD method on a single GPU is 0.37 s, while the iteration time for traditional distributed training using 2 GPUs on a single node is 0.57 s, resulting in a performance improvement of approximately 1.52×. For larger models, such as G-10B, the iteration time of the BCD method on 8 GPUs is 2.41 s, while the iteration times for traditional distributed configurations with 16 GPUs across 4 nodes and 16 GPUs across 2 nodes are 3.17 s and 3.25 s, respectively, showing improvements of 1.32× and 1.35×. For even larger models, such as G-20B, the iteration time of the BCD method on 16 GPUs distributed over 2 nodes is approximately 1.76× faster than that of the traditional distributed setup using 32 GPUs over 4 nodes.

Across all experimental results, the BCD method achieves iteration time improvements ranging from 1.52× to 1.76×, with an average speedup of approximately 1.41×. These findings indicate that BCD not only delivers strong performance on smaller models but also offers substantial advantages in distributed training for larger models, requiring fewer GPU hours per iteration.

\begin{figure}[!ht]
    \centering
    \includegraphics[width=0.45\textwidth]{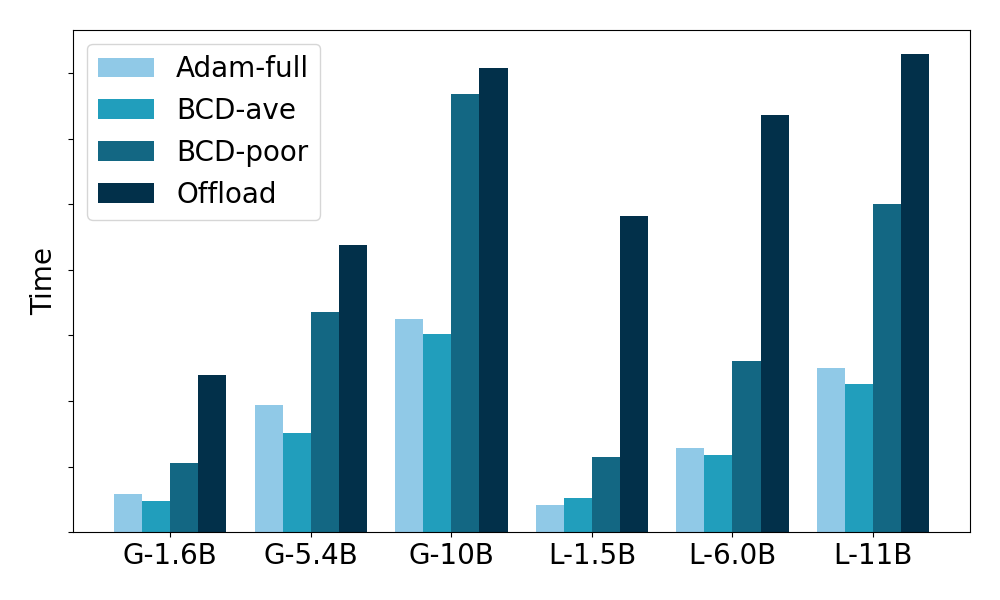}
    \caption{The training time of GPT and LLaMA models using different methods at various model scales.}
    \label{full training time}
\end{figure}

\subsubsection{Full training analysis}
The total training time can be calculated by multiplying the number of iterations by the iteration time per round. Considering that full-parameter training for 10B and 20B models requires an economic cost close to \$100,000, which is difficult to bear, we estimate the overall training time using Adam's average B-F multiplier and the worst B-F multiplier. As described in the B-F multiplier section, Adam's average B-F multiplier is 1.39. Therefore, for BCD-Adam, the ratio of the one iteration training time multiplied by 1.39 to the one iteration training time of the full-parameter Adam training method (including offloading and distributed methods) can be considered as the ratio of the training time for BCD-based models to that for full-parameter models. The results are shown in Figure \ref{full training time}, BCD outperforms the offloading's  42.0\%   in training time cost with the same computation resources. BCD matches distributed training speed using just half of the resources. For the worst-case scenario, we replace the average B-F multiplier in the above calculation with the the worst B-F multiplier, which is 2.89. This leads to the conclusion that, in the worst-case scenario, the training  speed of BCD-Adam, using half computational resources, is {48.9\%} slower than distributed methods but more than  {112.5\%} faster than offloading methods using equivalent resources.

Based on all the experimental results, it can be observed that the BCD method achieves a performance improvement of 1.32× to 1.76× compared to traditional methods in terms of iteration time, demonstrating a clear advantage.

\section{Theoretical cost analysis} 

\begin{table}[!htbp]
\centering
\renewcommand{\arraystretch}{1.2}
\setlength{\tabcolsep}{6.0pt}
\scalebox{0.65}{
\begin{tabular}{c|cc|cc|cc|cc}
\hline
\multirow{2}{*}{\textbf{}} & \multicolumn{2}{c|}{\textbf{G-1.6B}} & \multicolumn{2}{c|}{\textbf{G-5.4B}} & \multicolumn{2}{c|}{\textbf{G-10B}} & \multicolumn{2}{c}{\textbf{G-20B}} \\ \cline{2-9}
& \textbf{N/G} & \textbf{Cost } & \textbf{N/G} & \textbf{Cost } 
& \textbf{N/G} & \textbf{Cost }
& \textbf{N/G} & \textbf{Cost } \\ \hline

 \scalebox{1}{Full Update} &1/2 &{--}   &1/8 &{--}  &2/8 &{--}   &4/8 &{--}  \\ \hline
\scalebox{1}{BCD(The worst)} &1/1 &\textbf{-54.3\%}   &1/4 &\textbf{-28.9\%}  &1/8 &+2.6\%  &2/8 &\textbf{-28.2\%}  \\ \hline
\scalebox{1}{BCD(Average)}  &1/1 &\textbf{-80.4\%}   &1/4 &\textbf{-68.5\%} &1/8 &\textbf{-53.6\%}  &2/8 &\textbf{-68.3\%} \\ \hline
\end{tabular}}
\caption{The training cost reduction of BCD compared to full-parameter updates across different models on RTX 4090 cluster. N/G means the number of nodes and the number of GPU per node.}
\label{cost 4090}
\end{table}

\begin{table}[!ht]
\centering
\renewcommand{\arraystretch}{1.2}
\setlength{\tabcolsep}{6.0pt}
\scalebox{0.65}{
\begin{tabular}{c|cc|cc|cc|cc}
\hline
\multirow{2}{*}{\textbf{}} & \multicolumn{2}{c|}{\textbf{G-1.6B}} & \multicolumn{2}{c|}{\textbf{G-5.4B}} & \multicolumn{2}{c|}{\textbf{G-10B}} & \multicolumn{2}{c}{\textbf{G-20B}} \\ \cline{2-9}
& \textbf{N/G} & \textbf{Cost } & \textbf{N/G} & \textbf{Cost } 
& \textbf{N/G} & \textbf{Cost }
& \textbf{N/G} & \textbf{Cost } \\ \hline

 \scalebox{1}{Full Update} &\textbf{1/1} &{--}    &\textbf{1/2} &{--}  &\textbf{1/4} &{--}   &\textbf{1/8} &{--}  \\ \hline
\scalebox{1}{BCD(The worst)} &1/1 &\textbf{-72.2\%}   &1/4 &\textbf{-43.8\%} &1/8 &\textbf{-42.6\%}  &2/8 &\textbf{-55.1\%} \\ \hline
\scalebox{1}{BCD(Average)}  &1/1 &\textbf{-88.8\%}  &1/4 &\textbf{-75.5\%} &1/8 &\textbf{-74.9\%}  &2/8 &\textbf{-80.8\%} \\ \hline

\end{tabular}}
\caption{The training cost reduction of BCD compared to full-parameter updates across different models on A100 cluster.}
\label{cost a100}
\end{table}

\begin{table}[!ht]
\centering
\renewcommand{\arraystretch}{1.2}
\setlength{\tabcolsep}{6.0pt}
\scalebox{0.65}{
\begin{tabular}{c|cc|cc|cc|cc}
\hline
\multirow{2}{*}{\textbf{}} & \multicolumn{2}{c|}{\textbf{G-1.6B}} & \multicolumn{2}{c|}{\textbf{G-5.4B}} & \multicolumn{2}{c|}{\textbf{G-10B}} & \multicolumn{2}{c}{\textbf{G-20B}} \\ \cline{2-9}
& \textbf{N/G} & \textbf{Cost } & \textbf{N/G} & \textbf{Cost } 
& \textbf{N/G} & \textbf{Cost }
& \textbf{N/G} & \textbf{Cost } \\ \hline

\scalebox{1}{Full Update} &\textbf{1/1} &{--}    &\textbf{1/2} &{--}  &\textbf{1/4} &{--}  &\textbf{1/8} &{--}  \\ \hline
\scalebox{1}{BCD(The worst)} &1/1 &\textbf{-58.7\%}   &1/4 &\textbf{-9.5\%} &1/8 &\textbf{-6.7\%}  &2/8 &\textbf{-27.4\%} \\ \hline
\scalebox{1}{BCD(Average)}  &1/1 &\textbf{-82.5\%}   &1/4 &\textbf{-59.4\%} &1/8 &\textbf{-58.1\%}  &2/8 &\textbf{-67.7\%} \\ \hline
\end{tabular}}
\caption{The training cost reduction of BCD compared to full-parameter updates across different models on A800 cluster.}
\label{cost a800}
\end{table}

We evaluate the training cost of the BCD method compared to full-parameter updates across three hardware platforms: RTX 4090, A100, and A800. The analysis covers models ranging from 1.6B to 20B in parameter size, and the results are reported as the percentage reduction in total training cost. Both the average-case and worst-case B-F multipliers are considered. Detailed results are shown in Table~\ref{cost 4090}, \ref{cost a100}, and \ref{cost a800}.

Since RTX 4090 and A800 GPUs are interconnected using PCIe, which has limited bandwidth, the communication cost increases rapidly as the model size grows. Therefore, the BCD method, which requires less bandwidth, results in a larger economic cost reduction in larger model training cases. In the 1.6B cases, half of the GPU memory of A100 and A800 remains unused, leading to a waste of resources. As a result, the BCD method achieves a significant economic cost reduction.  

\section{Code and dataset links\label{code and Dataset}}
\begin{links}
  \link{Code} {https://github.com/quorvath/gpt2-megatron-deepspeed}
  \link{Datasets}{https://modelscope.cn/profile/quorvath}
\end{links}

\end{document}